\documentclass[5p,times]{elsarticle} 

\journal{Neurocomputing}

\usepackage{amssymb}

\usepackage[figuresright]{rotating}
\usepackage{svg}
\usepackage{hyperref}
\usepackage{booktabs}
\usepackage{amsmath}
\usepackage{color, colortbl}
\definecolor{light-gray}{gray}{0.95}
\newcommand{\code}[1]{\colorbox{light-gray}{\small\texttt{#1}}}

\usepackage[inline]{enumitem}

\begin{document}

\begin{frontmatter}

\title{Study of the performance and scalability of federated learning for medical imaging with intermittent clients}

\author[IFCA]{Judith Sáinz-Pardo Díaz\corref{correspondingauthor}}
\ead{sainzpardo@ifca.unican.es}
\author[IFCA]{Álvaro López García}
\ead{aloga@ifca.unican.es}

\address[IFCA]{Instituto de Física de Cantabria (IFCA), CSIC-UC \\ Avda. los Castros s/n. 39005 - Santander (Spain)}
\cortext[correspondingauthor]{Corresponding author}

\begin{abstract}
Federated learning is a data decentralization privacy-preserving technique used to perform machine or deep learning in a secure way. In this paper we present theoretical aspects about federated learning, such as the presentation of an aggregation operator, different types of federated learning, and issues to be taken into account in relation to the distribution of data from the clients, together with the exhaustive analysis of a use case where the number of clients varies. Specifically, a use case of medical image analysis is proposed, using chest X-Ray images obtained from an open data repository. In addition to the advantages related to privacy, improvements in predictions (in terms of accuracy, loss and area under the curve) and reduction of execution times will be studied with respect to the classical case (the centralized approach). Different clients will be simulated from the training data, selected in an unbalanced manner. The results of considering three or ten clients are exposed and compared between them and against the centralized case. Two different problems related to intermittent clients are discussed, together with two approaches to be followed for each of them. Specifically, this type of problems may occur because in a real scenario some clients may leave the training, and others enter it, and on the other hand because of client technical or connectivity problems. Finally, improvements and future work in the field are proposed.
\end{abstract}


\end{frontmatter}

\section{Introduction}

Recently, digitization and globalization have led companies and institutions of many different kinds to collect a vast amount of data. This can range from industrial and banking data to medical data. The large volume of information collected on a daily basis requires thorough analysis in order to infer new knowledge or make predictions, which has led to many advances in machine learning and deep learning techniques. However, the great value of such information also implies an imperative need to ensure the security and privacy of the analyzed data. The need to protect data privacy is in many cases a mandate of regulatory requirements (e.g., GDPR) \cite{yang2020federated}. 

The classic approach of applying machine learning to decentralized data (i.e. with several data owners) in a centralized way consist of sending the data from each and every one of the clients or data owners to a central server, training the model that returns the prediction, and then sending the prediction back to the corresponding client in each case.

\begin{figure}[ht]
    \centering
    \includegraphics[width = \linewidth]{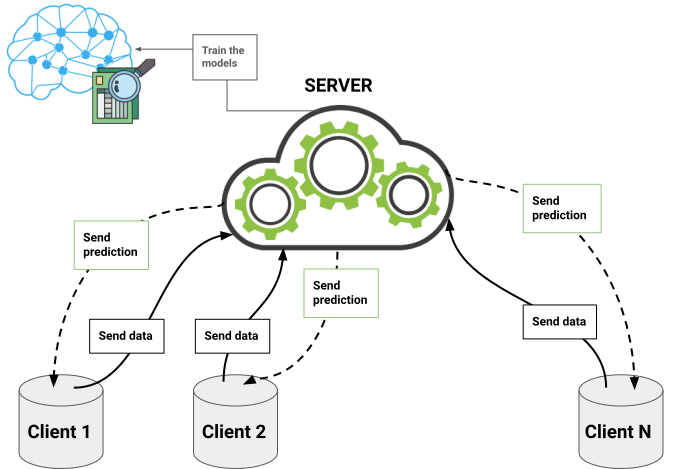}
    \caption{Diagram of the machine learning centralized approach.}
    \label{fig:cent_approach}
\end{figure}

This very first approach poses several disadvantages. The first and foremost is related with the security properties of the data itself: data must be sent from the data owners to the central server and therefore it can be intercepted. In addition to this, transferring those data require a network connection with enough bandwidth (when the data are large) and low latency (to return back the predictions in time).

One solution to some of these problems would be to collect all the initial data from the different clients, then the machine learning models are trained with this data, and finally a copy of the model is sent to each of the clients. In this case, the information does not have to travel every time new data is available, since the model is in the possession of each data owner. Compared to the classical approach, it has two essential advantages: the latency problem is eliminated (predictions are performed at the client), and the network dependence is reduced. However, the information still has to leave each client and travel to the central server for the initial training phase (or for any subsequent retraining).

In this case, even if we reduce the network dependency, data has to travel from the clients to the central server. As already mentioned, this presents a problem from the security point of view (i.e. sensitive data that can be intercepted, or data with privacy restrictions) but also from the technical point of view if large amounts of data are to be transferred, but also if the network quality is low (like in Internet of Things devices or in Edge computing cases). 

Federated learning (FL) is on the rise \cite{fl_systems} and comes handy when we are interested in ensuring that data does not leave the servers of each data owner, even for training purposes; or when due to privacy concerns about sensitive data, it is not feasible to collect all local data in the server's data center and perform centralized training. The main idea behind FL is to analyze data in a decentralized way, ensuring that user data will never be sent to central servers. This technique does not seek the security of the data through its anonymisation, but it seeks to achieve the analysis of the data without them leaving the device or the center that generates them.

Suppose we have two data owners that cannot share their data with each other, for example, two hospitals. However, they both want to do research in the same area, so collaboration between them is essential. Let us also suppose that in their research they need to use machine or deep learning models, for which there will be a mediator (the server) between them that will create and choose such models (for example, neural networks) to obtain insights from their data. A concrete example can be found in the case of medical imaging research. Here federated learning can be applied in the following way:
\textit{Hospital 1} receives from the server the the model to be trained using its data. Once it has done so, it sends back to the server the weights obtained after the training. \textit{Hospital 2} does the same tasks. With the weights obtained in each case, the server calculates some new aggregated weights. Once these new aggregated weights are available, they are sent back to both hospitals, which re-evaluate the model with these weights with their data. This process is repeated as many times as needed.

In this paper we will study the applicability of a federated learning schema and the performance obtained when varying the number of clients. Also two problems related to intermittent clients are analyzed, together with two possible approaches to be considered in each of them. In order to do so, we will implement a simple federated learning system that we will apply to a use case of medical image data. The idea of presenting this medical imaging use case arises from the motivation that it could be applied to a real case where different hospitals or research centers have patient data that cannot be shared among them, but could be key to improve their research results. By using an federated learning scheme, they could collaborate without sharing their data (which could contain sensitive information from different patients) either among themselves or with a central server, thus facilitating collaboration in several research areas. Prior to this analysis, we present federated learning background to serve as a basis for introducing into this field.
The remainder of this work is structured as follows:
In Section~\ref{sec:federated} we will introduce the federated learning technique and describe its main aspects, together with communication issues, types of Federated Learning, foundations on the aggregation functions among other issues. 
In Section~\ref{sec:related} we present the related work in this area. 
Afterwards, in Section~\ref{sec:use-case} we study the implementation and use of a federated learning system to carry out the data analysis described in 
this section, together with a comprehensive analysis of the use case when varying the number of clients and two problems related to intermittent clients. 
Finally, Section~\ref{sec:conclusions} draws the conclusions and future work for this study.

\section{Federated learning: background}
\label{sec:federated}
Artificial intelligence, machine and deep learning empower a wide range of applications (from anomaly detection to image classification or natural language processing among many others). In order to produce such systems, large amounts of labeled data are needed to build a robust application with an acceptable level of accuracy. These data need to be centralized or aggregated into a single location, in order to be able to consume it to build the model or applications. In some cases it is difficult or impossible to obtain a good quality and large enough dataset. This is the case of distributed datasets which cannot be centralized into a single location, due to different reasons, such us technical limitations or even privacy concerns. 

The \emph{federated learning} term (FL) was firstly introduced by McMahan in 2017 \cite{comm_decen_data} and refers to a technique that allows to build data driven models exploiting distributed data without the need to centrally store it. In a FL scenario, the learning task of a model is shared across a loose federation of different users or devices (called clients) that are coordinated by a central server. Each client is the data owner of a local training dataset that is used to compute an update to the current global model maintained by the server. Therefore, client data is never uploaded to the server, and only the updated model is shared. Federated learning brings a collaborative and decentralized approach to machine and deep learning, enabling the creation of smarter and more complete models, solving the latency problem and ensuring user privacy and has gained a lot of interest over the last years. 

In a simple cycle of a federated learning process, each of the clients trains the same model (provided by the server) with its own data. Once the training is completed, each client transmits the weights or parameters obtained to a central server. This server receives the weights calculated by each client, ideally through an encrypted channel, and aggregates them. To do this, an aggregation function is used, which in the simplest configuration can be an arithmetic mean. Finally, the server updates the model with the new aggregated weights and send it back to the clients. This concludes the first training cycle, which is repeated as many times as necessary (a number of rounds $N_{r}$ can be fixed).

\begin{figure}[ht]
    \centering
    \includegraphics[width = \linewidth]{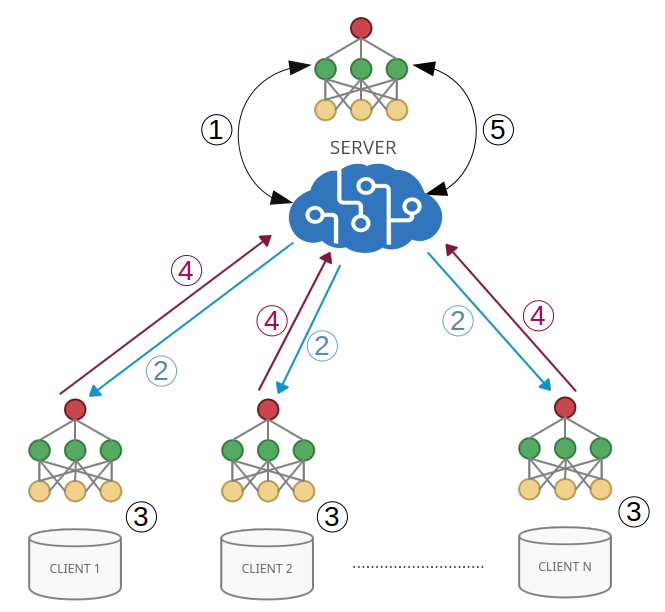}
    \caption{Scheme of a federated learning approach.}
    \label{fig:Decentralized_FL}
\end{figure}

The Figure~\ref{fig:Decentralized_FL} shows a schematic view of federated learning. The procedure  to be followed to train the models using federated learning (see \cite{fl_dp_softwaretools}) is described in what follows: 

\begin{enumerate}
    \item \textbf{SERVER}: creates the model to be trained locally by each client. To do this we can use the Python libraries \textit{tensorflow}, \textit{keras}, \textit{PyTorch}, \textit{sklearn}, etc.
    
    \item \textbf{SERVER}: transmits the model to the clients.
    
    \item \textbf{CLIENT}: each of them trains the model with its local data. We will leave part of the initial data for testing, in order to test the accuracy of the model locally on unseen data. It is also convenient to leave a part of the data we trained with as validation during and after training, as in any other learning process.
    
    \item \textbf{CLIENT}: each of them sends the local parameters to the server. Here it is important to note that in no case is raw data transmitted, only the parameters that define the model. Likewise, this must be done in an encrypted form, as information about the data could be extracted from it. 
    
    \item \textbf{SERVER}: aggregates the weights of each client using an aggregation operator and updates the model.
    
    \item Repeats the process from step 2.
\end{enumerate}

There are challenging aspects of FL, when compared with a traditional ML algorithm, that exist due to the distributed nature of the technique and can be grouped as follows  \cite{li2020federated}:
\begin{enumerate*}
    \item expensive communication, 
    \item systems heterogeneity, 
    \item statistical heterogeneity, and 
    \item privacy concerns.
\end{enumerate*}

\subsection{Communication issues}

Communication ---both in terms of privacy and performance--- between the server and the client is the most critical point of the process. One the one hand, each time the federated learning process is repeated, the client sends a new update of the calculated weights, and the difference between the updated weights and the previous can cause information to be extracted from the initial data. It is therefore crucial that this communication is encrypted to ensure its privacy and security. In addition, Differential Privacy (DP) or Homomorphic Encryption (HE) techniques can also be applied to ensure privacy \cite{FL_with_DP, FL_with_HE}. On the other hand, network performance can be a bottleneck when the model updates are large, therefore a compression or reduction of the data transferred back to the server should be performed in order to save bandwidth or to maximize the chances of a successful transfer when the network quality is low. In this regard, techniques such as data compression, dimensionality reduction or other techniques \cite{doi:10.1073/pnas.2024789118} are used to implement a more efficient communication method.

In the case of classification of medical imaging that will be studied in this paper, we will not focus on the cost of communication, since our main interest is to analyze the evolution of accuracy, loss and AUC as a function of the number of clients and the number of rounds performed (and therefore we will see the variation of the training time), as well as the study of the problem of intermittent clients. Regardless of the computational capacity of each client (directly related to the training time), the results in terms of these metrics will be the same. Therefore, as the analyzed use case is a simulated example, the same CPU will be considered for the training of all clients.

\subsection{Systems heterogeneity}

Another important challenge, caused by the underlying systems heterogeneity, is due to the fact that devices can be intermittent through the learning cycle. Let us see it with a classic example of federated learning application, the case of predictive keyboards. It could happen that one of the nodes that is used to train the models, a smartphone, is inactive at certain moment and cannot be used \cite{OpenProblemsFLSL}. As we have briefly described in the case of hospitals, collaboration between different data owners can be key, as individual data owners may not have a significant volume of data to analyze. In relation to the case study that will be analyzed in this paper, we can assume that different hospitals want to collaborate to classify X-Ray images, without sharing the patient's image data. For this task, it is proposed to apply a FL schema, but the number of clients (hospitals) may initially be limited, and as the study progresses, other hospitals or centers may decide to join the collaboration, while others may decide to leave it, e.g. because no new data is available or even for privacy concerns. In addition, due to connectivity issues or problems with computing infrastructures, some hospitals may not send their updates on time, which is another example of intermittent clients that will be analyzed.

\subsection{Non independent and identically distributed (non-i.i.d.) data vs heterogeneous data}

A very significant aspect that can affect the training process of a federated learning scheme is the distribution of the data among the different clients. In particular, the fact that these may be non-i.i.d. is a point to be taken into account. Specifically, in the Horizontal FL approach, we usually refer to non-i.i.d. data in terms of the skew of the different labels. For example, in extreme cases where there are clients that do not have all possible categories represented or only have one of them, it may happen that the global model fails to converge. In particular, this especially affects parametric models (e.g. neural networks) in the case of horizontal FL. These problems can be notably seen if it is the case that the distributions of the clients are too far away from the distribution of all the data globally. For example, the \textit{FedAvg} method, which consists of averaging the weights (or the models obtained when training locally in each client), is highly sensitive in the case of non-i.i.d. data \cite{zhu2021federated}.

However, as presented in \cite{OpenProblemsFLSL}, there are other approaches that can be studied in order to analyze whether the data are non-i.i.d. or homogeneous (again, as far as horizontal FL is concerned), not only the distribution of the labeling of the data. Although the intuitive idea is to analyze how different the $\mathcal{P}_{i}$ and $\mathcal{P}_{j}$ distributions of the labels in clients $i$ and $j$ are, other options that can make the data non-homogeneous are: imbalance in the number of data (clients with much more data than others), differences in the distribution of the features, or the fact that very different features have the same label. 

\subsection{Types of federated learning}\label{sec:types}

It is important to note that there are multiple types of federated learning \cite{OpenProblemsFLSL}, for example, we will start by distinguishing two variants according to the types of clients. Firstly, Cross-Device federated learning, where the clients are devices such as, for example, smartphones, or Cross-Silo federated learning, where the clients are different institutions (e.g. banks, hospitals, companies...), allowing for example that in the latter case the computing resources of each of them can be more powerful and scalable. 
 
On the other hand, two types of federated learning can also be distinguished depending on how the data is introduced: horizontal and vertical federated learning (also called homogeneous and heterogeneous FL respectively).
 
\begin{figure}[ht]
     \centering
     \includegraphics[width = \linewidth]{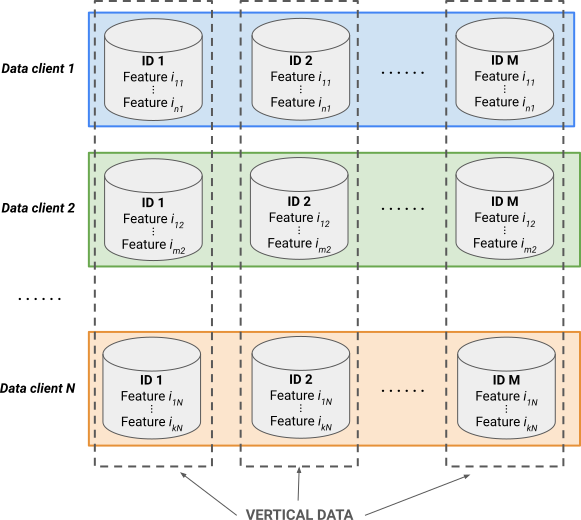}
     \caption{Example of vertical data.}
     \label{fig:vertical_data}
 \end{figure}

 \begin{itemize}
     \item \textbf{Horizontal federated learning}. This is the most intuitive case, and specifically the one that will be developed during the image classification use case presented in this study. It consists of considering the data of all clients with the same features. That is, although each client will have different data (different samples), they will all have the same features, for example, in the case of structured data, they will all have the same columns. Note that this is the type of FL that will be applied in the case study of this work.
     
     \item \textbf{Vertical federated learning}. In contrast to the previous case, now the different clients have data with different characteristics, but with the same identifier. That is, suppose we have two institutions (each of them will be a client of the FL schema), which have data from the same $n$ users, but each of them has information about different characteristics. Then, the data will be vertically condensed considering the information held by the two institutions for user $i$. A priori it is a less intuitive and more complex approach, but it is very useful in many cases in the cross-silo federated learning cases. A use example of this case can be seen in \cite{FederatedLearning_Vertically}. To better understand this approach, an example is presented in Figure~\ref{fig:vertical_data}, assuming $N$ clients, the first one with $n$ features, the second one with $m$, and client $N$ with $k$ features. Moreover, all of them with data corresponding to the same $M$ IDs. 
 \end{itemize}
 
Finally, highlight the concept of gossip learning as a variant of the classic federated learning architecture in which there is no dependence on a central server. The latter is of particular interest in cases where clients do not want to depend on a third party to carry out the process, and have sufficient resources to carry out the process of aggregating weights and updating the model themselves with a consensus algorithm.

\subsection{Aggregation function}\label{sec:aggregation}

When the model is trained locally by the different clients and once the weights or parameters calculated after such training are obtained, the server must aggregate them to update the model. Different functions can be considered in order to make this aggregation \cite{fl_dp_softwaretools, wang2020federated, PFedAvg, PALIHAWADANA2022432}. One of the most commonly used method for aggregation is known as \textit{federated averaging}. As its name indicates, it simply consists of taking the means of the weights (or that of the models) calculated with each client. Thus, the results obtained with this mean (p.e. in the case of a neural network, the average of the weights obtained in each layer is calculated) will be the new weights of the model, from which the next iteration of the federated learning process will begin. 

In our study we will use the \textit{federated weighted averaging} as the aggregation function. In particular, this operator is similar to the classic federated averaging operator, but it performs the average on a pondered way in order to take into account the possible imbalance according to the number of data of each client.

Then, be $N$ the number of clients, $n_{i}$ the number of data of client $i$ $\forall i \in \{1,\hdots,N\}$. We define $w_i$ as follows:

\begin{equation}
w_i = \frac{n_i}{\sum_{i=1}^{N}n_{i}} \hspace{0.5cm} \forall i \in \{1,\hdots,N\}
\label{wi}
\end{equation}

Let $x_{i}$ be the weights obtained for the model after training it with the data of client $i$ $\forall i \in \{1,\hdots,N\}$. The aggregation of the weights ($x_{aggregated}$) is calculated as follows:

\begin{equation}
x_{aggregated} = \sum_{i=1}^{N}w_{i} x_{i} 
\label{x_agg}
\end{equation}

Thus, once the server has calculated $x_{aggregated}$, it can update the model to send it back again to each client and repeat the process as many times as it deems necessary.

Although in this work we will use the federated weighted averaging operator exposed previously, it is also interesting to study the application of other aggregation functions. In particular using machine learning models in order to optimize weight aggregation is really attractive.

As mentioned previously, additional privacy conditions may be applied to the weights before aggregation, in order to limit the loss of information allowed if they are intercepted during the client-server communication. For example, among other measures, it is possible to apply Differential Privacy in order to add noise to the weights. To better understand this case, suppose we have a list of numeric data, to which we are going to add noise by adding a random number between $-a$ and $a$ from a uniform distribution, with $a \in \mathbb{R}^{*}_{+}$. Then, by the \textit{Law of Large Numbers} it can be observed that when the number of data to which such noise is added tends to infinity, the mean of such noise will tend to zero (and therefore the mean with noise will tend to the mean of the original data). Specifically, note that the mean of a uniform distribution is given by $\mu = \frac{x+y}{2}$, and in this case $x=-a$ and $y = a$, so $\mu = 0$. Be $\bar{X}_n$ the mean of the $n$ random numbers, by the \textit{Strong Law of the Large Numbers}, $P(lim_{n \to \infty} \bar{X}_{n} = \mu) = 1$. Then, the mean of the weights with the added noise will tend to the mean of the original weights, as the mean of the error will tend to $\mu = 0$.

As already mentioned, regardless of the possibility of applying this type of privacy preservation measures, it is important to note that the weights must be transmitted to the server via encrypted communication.

\section{Related work} 
\label{sec:related}

Since the introduction of the \emph{federated learning} term by McMahan in 2017 \cite{comm_decen_data} there has been significant work on the topic. 
Federated learning is being studied in a wide range of fields \cite{li2020review} such as medical data \cite{rieke2020future,brisimi2018federated,xu2021federated,info:doi/10.2196/20891,antunes2022federated}, cybersecurity \cite{alazab2021federated,ghimire2022recent} or Internet of Things \cite{savazzi2020federated, imteaj2021survey} among others.
Due to the increased concerns,  restrictions and requirements related to data privacy there are numerous papers that expose different FL techniques, definitions and methodologies \cite{fl_systems, wang2020federated, jiang2019improving, FederatedLearning_Vertically}, as well as some of the problems that are still open \cite{OpenProblemsFLSL}, like the application of the technique in training data are non-i.d.d. on the local clients \cite{zhu2021federated} or the efficient transmission of the model parameters \cite{ZHU2021309}.

In addition to scientific literature, there are various software implementations providing federated learning tools, techniques and algorithms. The following is a brief presentation of some of the most prominent Python libraries for FL tasks (among others): \textit{Tensorflow Federated (TFF)} \cite{TFF} is focused on experimental and research use of federated learning techniques and data decentralization in general. \textit{PySyft} \cite{PySyft}, created by the OpenMined community, seeks to apply deep and machine learning techniques securely and privately. \textit{IBM federated learning} \cite{IBM_FL} is a Python framework for federated learning applied to a business environment. \textit{Flower} \cite{beutel2020flower} is a novel federated learning framework that is agnostic to the toolkit being used by the client devices. \textit{FedML} \cite{chaoyanghe2020fedml} is an open research library that seeks to facilitate the development of new federated learning algorithms. In addition to these, there are numerous Python-based libraries that implement Federated learning techniques, for example, in \cite{fl_dp_softwaretools} different use cases of the framework \textit{Sherpa.ai FL} are presented together with indications about the code implemented and theoretical aspects of the field. However, as will be explained in the following section, in our work we have decided to implement independently from these libraries the client and server classes necessary to carry out the classical federated learning scheme. 

Regarding the aggregation functions, as discussed in Subsection \ref{sec:aggregation} the most conventional approach is to use federated averaging (FedAvg), or even its weighted version as will be done in this study. However, there are other approaches that can be considered, such as CO-OP \cite{fl_dp_softwaretools} or federated matched averaging (FedMA) \cite{wang2020federated} among others.

In this paper, part of the case study exposed in Section~\ref{sec:use-case} has focused on the problem of intermittent clients, therefore, it is of great interest to review the state of the art regarding the selection of clients in a FL scheme, as well as possible problems with them. In \cite{https://doi.org/10.48550/arxiv.2010.13723} a novel client subsampling scheme in proposed in order to address the problems related with the client-server communication. In \cite{8761315}, FedCS is proposed as an efficient FL protocol that actively manages clients according to their resources. Finally, in \cite{9443523} a communication-efficient client selection strategy is proposed to deal with communication limitations and intermittent client availability.

On the other hand, it is worth mentioning two variants of the classic federated learning schema, gossip learning and split learning \cite{vepakomma2018split, FL_IEEE}. The case of gossip learning (already mentioned in Subsection~\ref{sec:types}) is very intuitive once the concept of federated learning is understood, since it consists of eliminating the dependence on a central server to orchestrate the process. Furthermore, in the gossip architecture, the different clients (or nodes) randomly or conditionally select multiple clients to exchange updates between them at each repetition of the cycle \cite{Su2022}. Finally, in the simplest split learning configuration, each client trains a neural network up to a cut-off layer, and the output is sent to the central server. Five different federated learning architectures are presented in Figure 2 of \cite{Su2022}, being the Parameter Server (PS) architecture the one which will be implemented in this study.

Finally, as far as the implementation of a FL scheme and use case examples are concerned, we highlight the following works: \cite{info:doi/10.2196/20891}, where three benchmark datasets are studied (MNIST, Medical Information Mart for Intensive Care-III dataset and an ECG dataset), \cite{li2019privacy}, which shows experiments using MRI scans and include the use of Differential Privacy (DP) techniques, and \cite{10.1145/3501296}, which is a survey on the use of FL in smart healthcare. Compared to the previous ones, in addition to exhaustively presenting a use case, a comparison based on the number of clients, in order to analyze scalability is carried out, along with the exploration of two different approaches for the scenario in which there are intermittent clients (one disappears and a new one enters), and two others for the case in which a certain client does not send the weights corresponding to the current repetition in time for aggregation, but sends them later during the course of another repetition of the FL schema. 

\section{Experiment: Chest X-Ray image classification}\label{sec:use-case}

The objective of this study is to present the implementation of a simple example of a FL scheme from scratch. Specifically, the experiments have been carried out using the Python 3 programming language \cite{van1995python}.
In particular, the following additional libraries and versions have been used (among others): 
\code{scikit-learn} (0.24.2),
\code{tensorflow} (2.6.0),
\code{keras} (2.6.0),
\code{pandas} (1.3.4) and
\code{numpy} (1.21.4). 
For more information about this libraries see \cite{nguyen2019machine}. 

Although several Python libraries that implement the federated learning architecture have been presented in Section~\ref{sec:related}, in this case, for the sake of completeness of the study and greater customization, as well as better understanding of the architecture, we have made our own implementation. Then, in order to reproduce the procedure exposed in Section~\ref{sec:federated}, and in particular in Figure~\ref{fig:Decentralized_FL}, we have implemented a class called \emph{Client} and another one called \emph{Server} to follow this scheme.

\subsection{Data used}

In this work we show a medical image analysis use case. Specifically, we are going to use chest X-Ray images, and the objective is to classify them according to whether or not the patient has pneumonia. Specifically, the data were obtained from \cite{Data_Xray}.

The data used (which are published openly) are divided into three groups: train, test and validation. Let us present in Table~\ref{tab:distribution_data} the distribution of the images of each type (pneumonia and normal) in each of these three sets.

\begin{table}[ht]
\centering 
\begin{tabular}{@{}lcc@{}}
\toprule
 & \textbf{Pneumonia} & \textbf{Normal}  \\
\midrule
        \textit{Train} & 3875 & 1341 \\
        \textit{Test} & 390 & 234 \\
        \textit{Validation} & 8 & 8\\
\bottomrule
\end{tabular}
\caption{Distribution of the X-Ray images in train, test and validation sets.}
\label{tab:distribution_data}
\end{table}

\begin{figure}[ht]
    \centering
    \includegraphics[width = \linewidth]{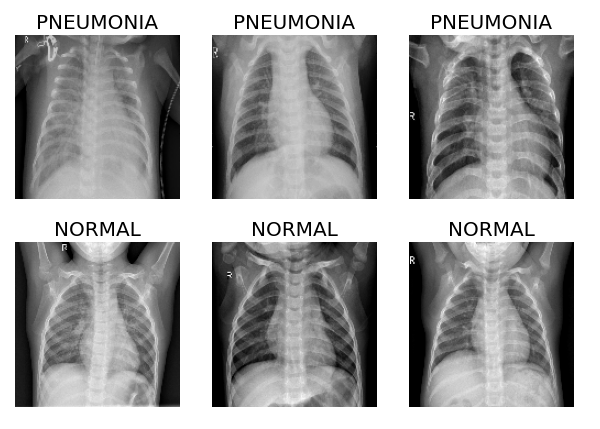}
    \caption{Example of the two categories of images under study (pneumonia and normal).}
    \label{fig:example_xray}
\end{figure}

To give an idea of the type of images we are working with, in Figure~\ref{fig:example_xray} three images labeled as pneumonia, and three others labeled as normal (the patient does not present pneumonia) are shown, both obtained from the train set.

\subsection{Model under study}\label{sec:model}
In order to classify the previously presented images into normal or pneumonia, a model consisting of a multi-layer convolutional network is proposed (remember that the objective of this example is to predict from an X-Ray image whether the patient has pneumonia or not). Specifically, its structure is based on that of models studied for the classification of medical images (see \cite{Kaggle_Xray}). In addition, different tests have been carried out to adjust the final model that has been used in this use case, regarding the number of layers and the number of neurons in them, the compilation method, etc. The architecture of the final model applied has been implemented using the Python library \code{keras}, and it is shown below: 
\begin{itemize}[leftmargin = 0cm, label = {}]
    \item \textbf{Conv2D layer}. Filters: 32. Kernel size: (3, 3). Activation: \textit{ReLU}. Input shape: (150,150,1).
    \item \textbf{BatchNormalization layer}.
    \item \textbf{MaxPooling2D layer}. Pool size: (2,2). Strides: 2.
    \item \textbf{Conv2D layer}. Filters: 64. Kernel size: (3, 3). Activation: \textit{ReLU}.
    \item \textbf{Dropout layer}. Rate: 0.1.
    \item \textbf{BatchNormalization layer}.
    \item \textbf{MaxPooling2D layer}. Pool size: (2,2). Strides: 2.
    \item \textbf{Conv2D layer}. Filters: 64. Kernel size: (3, 3). Activation: \textit{ReLU}.
    \item \textbf{BatchNormalization layer}.
    \item \textbf{MaxPooling2D layer}. Pool size: (2,2). Strides: 2.
    \item \textbf{Conv2D layer}. Filters: 128. Kernel size: (3, 3). Activation: \textit{ReLU}.
    \item \textbf{Dropout layer}. Rate: 0.2.
    \item \textbf{BatchNormalization layer}.
    \item \textbf{MaxPooling2D layer}. Pool size: (2,2). Strides: 2.
    \item \textbf{Conv2D layer}. Filters: 256. Kernel size: (3, 3). Activation: \textit{ReLU}.
    \item \textbf{Dropout layer}. Rate: 0.2.
    \item \textbf{BatchNormalization layer}.
    \item \textbf{MaxPooling2D layer}. Pool size: (2,2).
    \item \textbf{Flatten layer}.
    \item \textbf{Dense layer}. Units: 128. Activation: \textit{ReLU}.
    \item \textbf{Dropout layer}. Rate: 0.2.
    \item \textbf{Dense layer}. Units: 1. Activation: \textit{sigmoid}.
\end{itemize}

In addition, the optimizer \textit{RMSprop} and the binary cross-entropy as loss function were used to compile the model. The metric used to quantify the performance of the model was the accuracy (number of correct predictions out of total predictions).

Finally, as the data have been randomly distributed along the different clients, there may be cases of imbalance in such data. In order to compare the centralized and decentralized cases, and in the latter case the evolution in the initial test set as a function of the number of FL cycle repetitions, the area under the ROC curve (AUC) will also be computed, for consistency of such results, using Python library \code{sklearn}. 

\subsection{Federated learning approach (3 clients)}
In the classical use of federated learning, the number of clients is predefined. However, in our use case we will be interested to see if this technique can bring advantages over the use of centralized data (when privacy requirements allow it). Thus, given the train set of data presented previously, different clients will be simulated in order to compare these approaches, and thus the application of a federated learning schema will be exposed as well as the results obtained by applying it. Note that by the actual way in which the different clients are created, it is a case of horizontal federated learning. 

Given the initial set of data (the train set), we will first divide it into 3 clients, with a different number of data for each of them, as presented in Table~\ref{tab:tab1}. In addition, in order to train and test the models on each client, we split each of these into train (75$\%$) and test (25$\%$) sets. We also present in this table the average time per epoch that it takes to train the model in each client. It is important to highlight that even though for this use case the different clients have been artificially created (from a centralized dataset), in the case of a real federated learning scheme, the communication of the weights between the different clients and the server must be encrypted. This would not be necessary if we simply wanted to present a distributed machine learning scheme, where the central server distributes different subsets of data among different workers.
\begin{table}[ht]
    \centering
    \begin{tabular}{@{}cccc@{}}
         \toprule
         \multicolumn{1}{c}{ } &   \multicolumn{2}{c}{\textbf{Number of data}} & \\
          \cmidrule{2-3}
         \multicolumn{1}{c}{ } &  \textbf{Train} & \textbf{Test} & \textbf{Average time per epoch (s)}\\
         \midrule
         \textit{Client 1} & 1050 & 350 & 23.1\\
         \textit{Client 2} & 1800 & 600 & 40.1\\
         \textit{Client 3} & 1062 & 350& 24\\
         \bottomrule
    \end{tabular}
    \caption{Number of data of each client and average training time per epoch. Case: 3 clients.}
    \label{tab:tab1}
\end{table}

Be $N_{c}$ the number of clients, $N_{e}$ the number of epochs the model is trained on each client, $N_{r}$ the number of times the FL schema is repeated and $t_i$ the average time it takes to train each epoch for the data of client $i$. Since this training is done in parallel, the execution time will be:

\begin{equation}
    N_{r} N_{e} \max_{i\in \{1,\hdots,N_{c}\}}t_{i} 
\label{ex_time}
\end{equation}

Note that training this same model in a centralized way, that is on the initial data set, being these in total 5216 images, the average time per epoch (average over 10 epochs executed) is 138.6s. Thus, if the process is repeated 10 times for the decentralized case ($N_r = 10$), the average time considering that one epoch is trained in each case ($N_e = 1$), will be 401s (see Table~\ref{tab:tab1}), while if we train the model for the centralized case 10 epochs, the execution time will be approximately 1386s. We are therefore interested to see if this substantial saving in execution time will be reflected in an accuracy penalty for the test set.

Therefore, let's see in Table~\ref{tab:cen_dec_3clients} the results obtained when training these two cases for the test set, which consists of 624 images. Note that in the centralized approach we are training with more data than with the 3 clients together in the decentralized one. This is because for the decentralized case we have left a part of the data as a test for further analysis. Even so, the results for the test accuracy obtained are better in the case where we apply federated learning (i.e. the data is used in a decentralized way). This does not occur with the AUC, which is slightly worse in the case of FL, but at the cost of a runtime reduction of more than 70$\%$. In fact, in addition to improved results in terms of accuracy, a reduction in execution time of approximately 71.07$\%$ is obtained.

\begin{table}[ht]
    \centering
    \begin{tabular}{@{}ccc|ccc@{}}
    \toprule
    \multicolumn{3}{c|}{\textit{Centralized approach}} & \multicolumn{3}{c}{\textit{Decentralized approach}} \\
    \midrule
    \textbf{Loss} & \textbf{Accuracy} & \textbf{AUC}  & \textbf{Loss} & \textbf{Accuracy} & \textbf{AUC} \\
    \textbf{(test)} & \textbf{(test)} & \textbf{(test)} & \textbf{(test)} & \textbf{(test)} & \textbf{(test)}\\
    \midrule
    3.0694 & 0.6619 & 0.9429 & 2.6034 &  0.8029 & 0.9185\\
    \bottomrule
    \end{tabular}
    \caption{Centralized approach vs decentralized approach. Case: 3 clients.}
    \label{tab:cen_dec_3clients}
\end{table}

The evolution of the test loss, accuracy and AUC values obtained for this case, training one epoch each time on each client ($N_{e}=1$), and repeating this process $N_{r}$ times, is shown in Table~\ref{tab:tab2}.
\begin{table}[ht]
    \centering
    \begin{tabular}{@{}cccc@{}}
    \toprule
    $N_{r}$ & \textbf{Loss (test)} & \textbf{Accuracy (test)} & \textbf{AUC (test)} \\
    \midrule
    1 & 8.5823 & 0.6250 & 0.5105 \\
    2 & 8.2261 & 0.6250 & 0.5204 \\
    3 & 11.8544 & 0.6250 & 0.5987 \\
    4 & 13.6620 & 0.6250 & 0.8799 \\
    5 & 8.9979 & 0.6266 & 0.8936 \\
    6 & 12.2626 & 0.6298 & 0.9210 \\
    7 & 3.0013 & 0.7660 & 0.9271 \\
    8 & 4.3232 & 0.7308 & \textbf{0.9313} \\
    9 & 6.2835 & 0.6987 & 0.9246 \\
    10 & \textbf{2.6034} & \textbf{0.8029} & 0.9185 \\
    \bottomrule
    \end{tabular}
    \caption{Decentralized approach. Metrics obtained for the test data varying $N_r$ and with $N_e=1$. Case: 3 clients.}
    \label{tab:tab2}
\end{table}

It can be observed that, contrary to what might be intuitively expected, the best results in terms o loss and accuracy for the set of tests are obtained after 10 repetitions, while the best value for the AUC is reached with $N_r = 8$, being better than the one obtained after 10 repetitions. This can be due to the fact that in each repetition the weights of the models are adjusted according to the data of each client, and the data on which we are predicting have not been seen by any client. In the same way, the convergence of the model can be seen when increasing repetitions of the FL cycle, especially noticeable in the AUC, which goes from 0.5105 to 0.9185 in the last round, and 0.9313 in the best case.

In the same way, as we will observe in the following table, the same thing will happen with the test set of each client, since we are not interested in overfitting the models for a specific client, but rather that based on the data of each client, they generalize in the best possible way. In Table~\ref{tab:test_metrics_3clients} the results in terms of loss and accuracy for each client's test set evaluating using the model obtained after $N_r$ repetitions, are presented. 

\begin{table}[ht]
    \centering
    \resizebox{\linewidth}{!} {
    \begin{tabular}{@{}ccccccc@{}}
    \toprule
    \multicolumn{1}{c}{ } &  \multicolumn{2}{c}{\textit{Client 1 (test)}} &  \multicolumn{2}{c}{\textit{Client 2 (test)}} &  \multicolumn{2}{c}{\textit{Client 3 (test)}} \\
    \midrule
    $N_{r}$ & \textbf{Loss} & \textbf{Accuracy}  & \textbf{Loss} & \textbf{Accuracy}  & \textbf{Loss} & \textbf{Accuracy} \\
    \midrule
    1  & 6.4700 & 0.7143 & 3.7583 & 0.8333 & 8.6489 & 0.6186\\
    2  & 6.0401 & 0.7143 & 3.4773 & 0.8333 & 8.0885 & 0.6186\\
    3  & 8.4223 & 0.7143 & 4.8034 & 0.8333 & 11.2106 & 0.6186\\
    4  & 7.2142 & 0.7200 & 4.1675 & 0.8333 & 10.2194 & 0.6186\\
    5  & 2.9167 & 0.7513 & 1.4952 & 0.8467 & 4.1782 & 0.6441\\
    6  & 3.1232 & 0.8000 & 1.5306 & 0.8750 & 4.5520 & 0.7118\\
    7  & \textbf{0.1318} & \textbf{0.9714} & \textbf{0.0357} & 0.9883 & \textbf{0.2064} & \textbf{0.9548}\\
    8  & 0.2194 & 0.9657 & 0.0505 & \textbf{0.9917} & 0.3503 & 0.9463\\
    9  & 0.3861 & 0.9457 & 0.1813 & 0.9817 & 0.9088 & 0.8927\\
    10  & 0.3553 & 0.9543 & 0.0921 & 0.9783 & 0.2908 & \textbf{0.9548}\\
    \bottomrule
    \end{tabular}
    }
    \caption{Metrics obtained for the test set of each client varying $N_r$ with $N_e=1$ fixed. Case: 3 clients.}
    \label{tab:test_metrics_3clients}
\end{table}
    
Note that for client 1, the best results in terms of accuracy are obtained for $N_r=7$, for client 2 with $N_r=8$ and with $N_{r}=7$ and $N_r=10$ for client 3. In terms of the loss function, the best results are obtained with $N_r=7$. That is, in most cases the best results for the tests set of each client are reached for $N_r=7$. It is again observed that more repetitions do not necessarily lead to better results, since these are made by aggregating the different weights obtained for each client, which will vary with each repetition. In addition, the convergence of the method can be clearly seen from the sixth repetition.

On the other hand, if we take $N_e=10$ and $N_r = 1$, although according to the Equation~\ref{ex_time} the execution time is the same as in the case where $N_e=1$ and $N_r = 10$, the results are actually worse. Specifically, in this case we obtain for the test set a loss of 11.8120, and an accuracy of 0.6250, as opposed to the results obtained for $N_e=1$ and $N_r=10$ (see the last row of Table~\ref{tab:tab2}). Concretely, taking $N_e=10$ and $N_r = 1$, we obtain the same accuracy as for the case $N_e=N_r=1$, and a worse value for the loss (see the first row of Table~\ref{tab:tab2}). This tells us that in this specific use case is more appropriate to perform more repetitions of the federated learning scheme instead of training more epochs on each client. In addition, this may be because training more epochs on each client leads to overfitting on this client, worsening the accuracy for the test set. Intuitively this reinforces the idea of using federated learning to improve results in a decentralized data case.

\subsubsection{Data distribution}
Let us study the distribution of the two categories present in the data for each of the three clients (see Table~\ref{tab:distribution_3clients}). The objective is to analyze if the two classes are distributed in the same way in the different clients, or if, on the contrary, a different distribution is obtained in each one of them.

\begin{table}[ht]
    \centering
    \resizebox{\linewidth}{!} {
    \begin{tabular}{@{}cccc@{}}
        \toprule
         &  \textbf{Number of data} & \textbf{Normal ($\%$)} & \textbf{Pneumonia ($\%$)}\\
         \midrule
         \textit{Client 1} & 1400 & 28.57 & 71.43 \\
         \textit{Client 2} & 2400 & 16.67 & 83.33 \\
         \textit{Client 3} & 1416 & 38.21 & 61.79 \\
         \midrule
         \textit{ALL} & 5216 &  25.71 & 74.29 \\
         \bottomrule
    \end{tabular}
    }
    \caption{Data distribution for the three clients (without distinguishing train and test sets).}
    \label{tab:distribution_3clients}
\end{table}

These data have been selected manually, so that the distributions of the data in the different clients are different. For example, in client 2 less than 17$\%$ of the samples correspond to images of patients without pneumonia, while this percentage exceeds 38$\%$ (more than double) in the case of client 3. In addition, if we compare with the global distribution of the data of the three clients, we can see that between that of the third one and the global distribution, there is more than a 12$\%$ difference. 
The distribution of the data is one of the most important factors that can affect a federated learning algorithm, in particular, the fact that they are unbalanced (as in this case), is a key aspect to be evaluated. However, as we have seen in the above, despite the unbalanced data, we managed to obtain a substantial improvement by distributing the data in 3 clients, compared to the case in which they are centralized (passing the accuracy from 0.6619 in the centralized case to 0.7853 in the best performance of the federated learning approach). In the following cases, the process of allocating the data among the different clients will be a random process, so that these differences in the distributions will not be so clear. 

\subsection{Federated learning approach (10 clients)}

Now, let us repeat the previous analysis using a larger number of data owners. In this case, instead of decomposing the train set into 3 clients, we will decompose it into 10 of them. Again we will use the model presented in Subsection~\ref{sec:model}. Table~\ref{tab:n_data_10clients} shows the number of data for each client as well as the average computation time per epoch is shown.

\begin{table}[ht]
    \centering
    \begin{tabular}{@{}lccccc@{}}
         \toprule
         \multicolumn{1}{c}{ } &   \multicolumn{2}{c}{\textbf{Number of data}} & \\
          \cmidrule{2-3}
         \multicolumn{1}{c}{ } &  \textbf{Train} & \textbf{Test} & \textbf{Average time per epoch (s)}\\
         \midrule
         \textit{Client 1} &  442 & 148 & 10.1\\
         \textit{Client 2} &  412 & 138 & 9.7\\
         \textit{Client 3} &  258 & 87 & 6\\
         \textit{Client 4} &  326 & 109 & 7.9\\
         \textit{Client 5} &  378 & 127  & 9\\
         \textit{Client 6} &  393 & 132 & 9\\
         \textit{Client 7} &  356 & 119 & 8\\
         \textit{Client 8} &  468 & 157  & 11\\
         \textit{Client 9} &  412 & 138 & 9.8\\
         \textit{Client 10} &  462 & 154 & 11\\
         \bottomrule
    \end{tabular}
    \caption{Number of data of each client and average training time per epoch. Case: 10 clients.}
    \label{tab:n_data_10clients}
\end{table}

As we have already explained in the previous case, when training the model in a centralized way, the average time per epoch (average over 10 epochs executed) is 138.6s. In this case, with 10 clients, if we repeat the process 10 times the average time considering that one epoch is trained in each case ($N_e = 1$), will be 110s (see Equation~\ref{ex_time}), while if we train the model for the centralized case 10 epochs, the execution time will be approximately 1386s. Let's see in Table~\ref{tab:tab_10clients_centr} the results of training these two cases for the test set. 

\begin{table}[ht]
    \centering
    \begin{tabular}{@{}ccc|ccc@{}}
    \toprule
    \multicolumn{3}{c|}{\textit{Centralized approach}} & \multicolumn{3}{c}{\textit{Decentralized approach}} \\
    \midrule
    \textbf{Loss} & \textbf{Accuracy} & \textbf{AUC}  & \textbf{Loss} & \textbf{Accuracy} & \textbf{AUC} \\
    \textbf{(test)} & \textbf{(test)} & \textbf{(test)} & \textbf{(test)} & \textbf{(test)} & \textbf{(test)}\\
    \midrule
    3.0694 & 0.6619 & 0.9429 & 4.7813 & 0.7212 & 0.9095\\
    \bottomrule
    \end{tabular}
    \caption{Centralized approach vs decentralized approach. Case: 10 clients.}
    \label{tab:tab_10clients_centr}
\end{table}

We can thus verify (although we will study this in more detail in a later summary of results), that applying federated learning techniques provides substantial advantages over the centralized case. The advantages in terms of execution time are evident (since the different clients will perform their computations in parallel, and the number of data for each one will always be strictly smaller than in the centralized case), and as shown in Table~\ref{tab:tab_10clients_centr}, there is a very significant improvement also with respect to the accuracy (although the values obtained for the other two metrics are slightly worse). The most significant strength is that it can be observed that in this case, the execution time of the decentralized case reduce by 92.06$\%$ the time of the centralized approach, while the accuracy has also increased.

In Table~\ref{tab:Nr_10clients} we show in detail the test loss, accuracy and AUC values obtained by varying the number of repetitions of the federated learning process (i.e. varying $N_r$, with $N_{e}=1$). Specifically, in this case, the best results for the three metrics are obtained with the highest number of repetitions performed, $N_{r}=10$. Looking at the precision, we can see that it is not until the sixth round that the value of this metric starts to increase, going from 0.6250 during the first 5 repetitions, to 0.7340 in the ninth, and decreasing a little in the tenth to 0.7212. While in the case of 3 clients (see Table~\ref{tab:tab2}) the accuracy did not exceed 0.6250 until the 5th repetition (when reached 0.6266), in this case it occurs from the 7th repetition onwards (with an accuracy of 0.6635 en this round). This shows that in this second case the convergence is slower, probably because each client has a smaller number of data than in the first case. In this case it is very interesting to note that in terms of the three metrics under study, loss, accuracy and AUC, the best results are obtained for round 9 instead of round 10, which would imply even less execution time, and therefore a temporal reduction of approximately 92.85$\%$ compared to the centralized case.  

\begin{table}[ht]
    \centering
    \begin{tabular}{@{}cccc@{}}
    \toprule
    $N_{r}$ & \textbf{Loss (test)} & \textbf{Accuracy (test)} & \textbf{AUC (test)} \\
    \midrule
    1 & 14.3417 & 0.6250 & 0.4237 \\
    2 & 8.6290 & 0.6250 & 0.4328 \\
    3 & 6.7486 & 0.6250 & 0.4609 \\
    4 & 16.5803 & 0.6250 & 0.5406 \\
    5 & 12.1097 & 0.6250 & 0.6825 \\
    6 & 9.6882 & 0.6250 & 0.7758 \\
    7 & 5.7260 & 0.6635 & 0.8763 \\
    8 & 5.5559 & 0.6907 & 0.9055 \\
    9 & \textbf{3.8730} & \textbf{0.7340} & \textbf{0.9130} \\
    10 & 4.7813 & 0.7212 & 0.9095 \\
    \bottomrule
    \end{tabular}
    \caption{Decentralized approach. Metrics obtained for the test data varying $N_r$ with $N_e=1$ fixed. Case: 10 clients.}
    \label{tab:Nr_10clients}
\end{table}

Let us now see in Table~\ref{tab:dec_10clients} for which values of $N_r$ (remember that $N_{e}=1$) the optimal results (in terms of loss and accuracy) are obtained for each client's test set, and what these values are. Note that the optimum of each metric is not always obtained for the same value of $N_r$, as is the case of clients 3, 7 and 10.

\begin{table}[ht]
    \centering
    \begin{tabular}{@{}lccc@{}}
    \toprule
    \multicolumn{1}{c}{ } & \textbf{Loss} & \textbf{Accuracy} & $N_{r}$\\
    \multicolumn{1}{c}{ } & \textbf{(client test set)} & \textbf{(client test set)} & \\
    \midrule
    \textit{Client 1} & 0.2779 & 0.9662 & 9\\
    \textit{Client 2} & 0.0001 & 1.0000 & 9\\
    \textit{Client 3} & 0.1060 & 0.9655 & 10\\
    \textit{Client 4} & 0.0330 & 0.9908 & 9\\
    \textit{Client 5} & 0.0192 & 0.9843 & 9\\
    \textit{Client 6} & 0.0940 & 0.9772 & 10\\
    \textit{Client 7} & 0.3918 & 0.9412 & 9\\
                      & 0.4982 & 0.9664 & 10\\
    \textit{Client 8} & 0.2372 & 0.9618 & 9\\
    \textit{Client 9} & 0.0029 & 1.0000 & 10\\
    \textit{Client 10} & 0.2432 & 0.9740 & 9\\
    \bottomrule
    \end{tabular}
    \caption{Decentralized approach. Optimal values for each client's test set. Case: 10 clients.}
    \label{tab:dec_10clients}
\end{table}

In Table~\ref{tab:dec_10clients} it can be seen that both in terms of loss and accuracy really successful results are obtained for the test set of each client, with the worst result being Client 7, and the best being Clients 2 and 9 (both with an accuracy of 100$\%$, and Client 2 with a loss value slightly lower than Client 9).

Again, as in the case of 3 clients, when taking $N_e=10$ and $N_r=1$ the estimated execution time is the same as in the case where $N_e=1$ and $N_r=10$ (see Equation~\ref{ex_time}), but the results are actually worse. Specifically, in this case we obtain for the test set a loss of 6.5796, and an accuracy of 0.6250, while this values are 4.7813 and 0.7212 respectively for the case of $N_e=1$ and $N_r=10$ (see the last row of Table~\ref{tab:Nr_10clients}). 

\subsection{Intermittent clients}

In the following, two problems that may arise related to client intermittency will be discussed. As mentioned previously, the problem of intermittent clients can be due to a wide range of reasons. The most common ones can be communication limitations, connectivity problems or issues related to computing infrastructures. However, in this case concerning the health field, if we assume that this study could involve the collaboration of different hospitals or health or research centers, all of them with X-ray images from different patients, the intermittency could be due to a new center deciding to participate in the training, or others deciding to drop out for other reasons, such us privacy concerns. Therefore, it is important to analyze the problems that may arise or how they could be addressed.

Suppose now that a new client enters the architecture, and one of the clients which was participating in the training, leaves it. This fits with a real case where a data owner decides to leave the training, either voluntarily or involuntarily (e.g. problems with internet connectivity, technical issues, or even privacy concerns). In the case of medical imaging, this can be very common if it is a study in which different hospitals participate, and some leave the training (for any of the reasons stated previously, among others), and others join it once it is started. To exemplify this, first let us come back to the case of a federated learning schema composed by 3 clients. Let us suppose that a client leaves (we will study what happens when client 1, 2 and 3 leave), and a new client enters, client 4, which will build using the validation data. 

Note that the new client that we are going to add (client 4), only consists of 16 images, compared to the 1400, 1416 and 2400 of those used previously. Thus, we seek to test the influence of the balancing of the datasets used. We consider two different options:
\begin{itemize}
    \item \textbf{Approach 1:} When a client leaves, the weights obtained for that client in previous repetitions are not taken into account for subsequent repetitions of the training. The weights obtained for the new client are included in the aggregation. 
    
    \item \textbf{Approach 2:} When a client leaves, its last weights calculated are kept and are used in subsequent aggregations to update the model. Again, the weights obtained for the new client are included in the aggregation. 
\end{itemize}

The results obtained for the prediction on the test set with each approach by adding client 4 and removing each of the previous clients are presented in Table~\ref{tab:remove_client_3clients}.
    \begin{table}[ht]
        \centering
        \begin{tabular}{@{}ccccc@{}}
        \toprule
         & \multicolumn{2}{c}{\textbf{Approach 1 (test)}} & \multicolumn{2}{c}{\textbf{Approach 2 (test)}} \\
        \cmidrule{2-5}
        \textbf{Client removed} & \textbf{Loss} & \textbf{Accuracy} & \textbf{Loss} & \textbf{Accuracy} \\
        \midrule
        \textit{Client 1} & \textbf{2.4668} & \textbf{0.8092} & 2.8923 & 0.7949\\
        \textit{Client 2} & 4.4945 & 0.7452 & \textbf{3.4701} & \textbf{0.7821}\\
        \textit{Client 3} & \textbf{2.1746} & \textbf{0.8237} & 3.5976 & 0.7340\\
        \bottomrule
        \end{tabular}
        \caption{Results obtained by eliminating one client and adding a new one with the two approaches described above. Case: 3 clients.}
        \label{tab:remove_client_3clients}
    \end{table}
    
In this case we cannot select one approach as better than the other, since by eliminating the first client, the first approach is better in terms of accuracy and loss than the second approach. The same applies to the third client, the second approach is better than the first one. However, the opposite happens when eliminating the second client (the one that contained a larger number of data), in this case approach 2 produces better results than approach 1, both in terms of loss and accuracy.

Following this same line, the results of removing a client and adding a new one in the case of the federated learning schema with 10 clients are shown. Again in this case the client that we add, client 11, will have as data the images of the validation set (16 images). We consider the same two approaches as in the case of three clients In Table~\ref{tab:client11_2approaches} the results obtained for the test set when removing each one of the initial ten clients and adding client 11 are shown.

    \begin{table}[ht]
        \centering
        \begin{tabular}{@{}lcccc@{}}
        \toprule
        & \multicolumn{2}{c}{\textbf{Approach 1 (test)}} & \multicolumn{2}{c}{\textbf{Approach 2 (test)}} \\
        \cmidrule{2-5}
        \textbf{Client removed} & \textbf{Loss} & \textbf{Accuracy} & \textbf{Loss} & \textbf{Accuracy} \\
        \midrule
        \textit{Client 1} & \textbf{4.0000} & \textbf{0.7500} & 4.5387 & 0.7131\\
        \textit{Client 2} & \textbf{3.6074} & \textbf{0.7564} & 4.8299 & 0.7083\\
        \textit{Client 3} & \textbf{3.7131} & \textbf{0.7548} & 4.3448 & 0.7228\\
        \textit{Client 4} & \textbf{3.7189} & \textbf{0.7548} & 4.3951 & 0.7196\\
        \textit{Client 5} & \textbf{3.7096} & \textbf{0.7548} & 4.2605 & 0.7228\\
        \textit{Client 6} & 3.7111 & \textbf{0.7548} & \textbf{3.6957} & 0.7340\\
        \textit{Client 7} & \textbf{3.9376} & \textbf{0.7436} & 4.8988 & 0.7051\\
        \textit{Client 8} & \textbf{3.4686} & \textbf{0.7580} & 4.3015 & 0.7147\\
        \textit{Client 9} & \textbf{3.3949} & \textbf{0.7596} & 3.7850 & 0.7260\\
        \textit{Client 10} & \textbf{3.6860} & \textbf{0.7548} & 8.1755 & 0.6266\\
        \bottomrule
        \end{tabular}
        \caption{Results obtained by eliminating one client and adding a new one with the two approaches described above. Case: 10 clients.}
        \label{tab:client11_2approaches}
    \end{table}

In this case, it is clear that in the majority of cases (9 out of 10 for the loss, and 10 out of 10 for the accuracy), the best results are achieved with the first approach (something that did not seem to be clear when we study the case of 3 clients). That is, without using in further rounds of the FL schema the weights obtained in previous repetitions if one client leaves the training process. 

It should be noted that the selection of one criterion or another will depend on the type of data of each client, but in principle, it seems more convenient not to keep the last weights calculated for the eliminated client, rather than not taking them into account (for example, the frequency with which data is updated for each client is key in considering whether or not to discard such weights).

Finally, let us suppose now that one of the clients involved in the process does not send the updates after a pre-established time. As the rest cannot wait indefinitely, a decision must be taken in advance: consider the previous update available or not count on this client. In the same way, once this client sends its update after a certain number of repetitions of the FL schema, it will have to be decided whether it will be included or whether it should not be included until it sends the weights corresponding to the current round. Then, the following two approaches can be studied:
\begin{itemize}
    \item \textbf{Approach A:} Until new weights are received from the client, the last update received by the client is used. Once a new one is received, it is included regardless of the repetition of the process. 
    \item \textbf{Approach B:} If a client does not send its updates on time in a certain round, the parameters are not included again, nor are those calculated in other repetitions used. The weights of that client will only be included again when the client re-trains the initialized model with the weights of the corresponding repetition.
\end{itemize}
Let us illustrate this problem with the following example: suppose that once the model is applied for the fifth time to each client, client $i$ (for certain $i \in \{1,\hdots,N_{c}\}$) does not send its weights after the pre-established maximum waiting time prior to aggregation. Also, suppose that this client sends the updates at the end of repetition 10. The results obtained with the two approaches exposed for this example are shown for each client in Table~\ref{tab:appAB_3clients} for the case of three clients and in Table~\ref{tab:appAB_10clients} for the case of 10 clients.

\begin{table}[ht]
        \centering
        \begin{tabular}{@{}lcccc@{}}
        \toprule
        & \multicolumn{2}{c}{\textbf{Approach A (test)}} & \multicolumn{2}{c}{\textbf{Approach B (test)}} \\
        \cmidrule{2-5}
        \textbf{Client \textit{i}} & \textbf{Loss} & \textbf{Accuracy} & \textbf{Loss} & \textbf{Accuracy}\\
        \midrule
        \textit{Client 1} & 12.2229 & 0.6250 & \textbf{8.6411} & \textbf{0.6522}\\
        \textit{Client 2} & 6.9331 & 0.6298 & \textbf{3.5930} & \textbf{0.7516}\\
        \textit{Client 3} & \textbf{2.0380} & 0.7436 & 3.0532 & \textbf{0.7997}\\
        \bottomrule
        \end{tabular}
        \caption{Results obtained for the test set considering the last repetition of the FL schema and approaches A and B. Client $i$ is the intermittent client considered in this example. Case: 3 clients.}
        \label{tab:appAB_3clients}
    \end{table}

In this case, we find that approach B achieves better results than approach one in all three cases in terms of accuracy, and in 2 of the 3 cases in terms of loss. This is not really significant enough to decide in favor of one approach or the other, so in Table \ref{tab:appAB_10clients} the results for the 10-clients case are studied. 

\begin{table}[ht]
        \centering
        \begin{tabular}{@{}lcccc@{}}
        \toprule
        & \multicolumn{2}{c}{\textbf{Approach A (test)}} & \multicolumn{2}{c}{\textbf{Approach B (test)}} \\
        \cmidrule{2-5}
        \textbf{Client \textit{i}} & \textbf{Loss} & \textbf{Accuracy} & \textbf{Loss} & \textbf{Accuracy}\\
        \midrule
        \textit{Client 1} & \textbf{5.6262} & 0.6250 & 6.7738 & 0.6250\\
        \textit{Client 2} & \textbf{4.1949} & 0.6250 & 4.3756 & 0.6250\\
        \textit{Client 3} & \textbf{4.9695} & 0.6250 & 8.2096 & 0.6250\\
        \textit{Client 4} & 5.1496 & 0.6250 & \textbf{4.8699} & 0.6250\\
        \textit{Client 5} & \textbf{4.5158} & 0.6250 & 6.4545 & 0.6250\\
        \textit{Client 6} & \textbf{4.3505} & 0.6250 & 4.3655 & 0.6250\\
        \textit{Client 7} & \textbf{4.4714} & 0.6250 & 6.3264 & 0.6250\\
        \textit{Client 8} & \textbf{3.4311} & 0.6250 & 6.0426 & 0.6250\\
        \textit{Client 9} & 5.1880 & 0.6250 & \textbf{4.9086} & 0.6250\\
        \textit{Client 10} & 5.4487 & 0.6250 & \textbf{4.8596} & 0.6250\\
        \bottomrule
        \end{tabular}
        \caption{Results obtained for the test set considering the last repetition of the FL schema and approaches A and B. Client $i$ is the intermittent client considered in this example. Case: 10 clients.}
        \label{tab:appAB_10clients}
    \end{table}

In this case, comparing approaches A and B for 10 clients, we can see that the models fail to converge in terms of accuracy by the end of the FL scheme rounds. This is not entirely surprising, since we have already seen that in this case it took 7 repetitions for the model to start converging by increasing the accuracy to 0.6250, and in this case, the model undergoes the change in the configuration of the clients in round 5. However, it is possible to compare these two approaches in terms of loss. Thus, in this example we can see that in 7 out of 10 cases, better results are obtained with approach A than with approach B, i.e. if a client does not send its updates in the corresponding repetition, the previous ones will be used until new weights are received. This is the opposite of what happened in the case of 3 clients. However, what can be clearly observed here is that once the FL scheme undergoes a change, namely some weights are not received in time and/or sent with delay, more rounds will be needed to converge. Again, as already mentioned, it is important to study which approach is suitable depending on the data used, the characteristics of the clients (and the number of data each one owns) and the objective of the problem.

\subsection{Summary and comparison}

In this section we will summarize the different results obtained in the analysis of the previous case study.

First, in Table~\ref{tab:comparativa} the best performance obtained in each case under study in terms of loss, accuracy and AUC (depending the round of the FL schema $N_{r}$) and time reduction vs the centralized approach is shown. 
\begin{table*}[ht]
    \centering
    \begin{tabular}{@{}lcccccc@{}}
         \toprule
         & \textbf{Test Loss} & \textbf{Test Acc.} & \textbf{Test AUC} & $N_{r}$ & \textbf{Exc. time (s)} & \textbf{Time reduction vs cent. apr. ($\%$)}\\
         \midrule
         \textbf{\textit{Centralized approach}} & 3.0694 & 0.6619 & 0.9429 & --- & 1386 & ---\\
         \midrule
         \textbf{\textit{Decentralized approach}} &  & & &  &  & \\ 
         \textit{3 clients} & \textbf{2.6034} & \textbf{0.8029} & 0.9185 & 10 & 401 & 71.07\\
                            & 4.3232 & 0.7308 & \textbf{0.9313} & 8 & 320.8 & 76.85\\
         \cmidrule{2-7}
         \textit{10 clients} & 4.7813 & 0.7212 & 0.9095 & 10 & 110 & 92.06\\
                             & \textbf{3.8730} & \textbf{0.7340} & \textbf{0.9130} & 9 & 99 & 92.85\\
         \bottomrule
    \end{tabular}
    \caption{Comparison of the centralized approach with the two decentralized cases under study. Best performance in terms of loss, accuracy and AUC.}
    \label{tab:comparativa}
\end{table*}

In the case of 10 clients it is clearly observed that the results for the three metrics are better in the case where 9 repetitions of the FL schema are performed instead of 10, which also leads to a shorter computation time. With respect to the case of 3 clients, it is observed that in terms of loss and accuracy, the best results are obtained in the tenth repetition, but this is not true for the AUC, which reaches its best value in the eighth repetition, again implying less computation time.

\begin{figure}[ht]
    \centering
    \includegraphics[width = \linewidth]{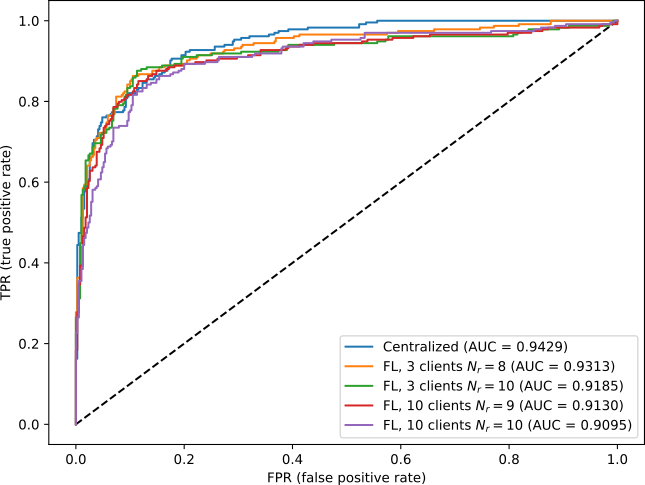}
    \caption{ROC curves for each of the cases exposed in Table~\ref{tab:comparativa}}
    \label{fig:roc}
\end{figure}

In addition, let's study it in more detail the AUC and the ROC curves obtained for the centralized approach and for the round which provides the best results in terms of AUC for the cases of three and ten clients, analyzing the one obtained for each round $N_{r}\in\{1,\hdots,10\}$, and for $N_{r}=10$ (see Figure \ref{fig:roc}). As already mentioned, it can be seen that the best results in terms of this metric are not obtained for the maximum number of rounds performed ($N_{r}=10$), but for 8 and 9 rounds respectively for the cases of 3 and 10 clients.

Looking again at the cases of three and ten clients, it is of particular interest to assess that this federated learning approach also gives good results in the test sets of each client. For this purpose in Table~\ref{tab:acc_loss_mean_clients} the average loss and accuracy obtained in each case is compared, considering the optimum value reached for each client (see Table~\ref{tab:test_metrics_3clients} and \ref{tab:dec_10clients}).

\begin{table}[ht]
    \centering
    \begin{tabular}{@{}lcc@{}}
         \toprule
         \multicolumn{1}{c}{ } & \textbf{Average loss} & \textbf{Average test accuracy}\\
         \multicolumn{1}{c}{ } & \textbf{(client test set)} & \textbf{(client test set)}\\
         \midrule 
         \textit{3 clients} & 0.1246 & 0.9726\\
         \textit{10 clients} & 0.1405 & 0.9786\\
         \bottomrule
    \end{tabular}
    \caption{Comparison of the average loss and accuracy obtained for the test sets of the clients of each decentralized case.}
    \label{tab:acc_loss_mean_clients}
\end{table}

In this case, in contrast to the previous one, it can be seen that the mean loss for the test sets of each client is lower when there are three clients than when there are ten (however, in the case of accuracy, the results are slightly better in the case of ten clients). 

In view of the above summary of results, it can be noted that in each case it must be studied whether a smaller or larger number of clients should be taken, depending on the characteristics of the data and the problem.

Now let us turn to the case study of the problem of intermittent clients. First, let us look at the average for loss and accuracy in the cases where a client is eliminated and a new one appears (Table~\ref{tab:remove_client_3clients_10clients}). Note that in both approaches the new client is the same (the one containing the images of the validation set). For this calculation the results of Tables~\ref{tab:remove_client_3clients} and \ref{tab:client11_2approaches} are used. In this case, the test set considered is the initial one (see Table~\ref{tab:distribution_data}), not that of each client.

    \begin{table}[ht]
        \centering
        \begin{tabular}{@{}lcccc@{}}
        \toprule
        \multicolumn{1}{c}{ } & \multicolumn{2}{c}{\textbf{Approach 1 (test)}} & \multicolumn{2}{c}{\textbf{Approach 2 (test)}} \\
        \cmidrule{2-5}
        \multicolumn{1}{c}{ } & \textbf{Average} & \textbf{Average} & \textbf{Average} & \textbf{Average} \\
        \multicolumn{1}{c}{ } & \textbf{loss} & \textbf{accuracy} & \textbf{loss} & \textbf{accuracy} \\
        \midrule
        \textit{3 clients} & 3.0453 & 0.7927 & 3.3200 & 0.7703\\
        \textit{10 clients} & 3.6947 & 0.7552 & 4.7223 & 0.7093\\
        \bottomrule
        \end{tabular}
        \caption{Results obtained by eliminating one client and adding a new one with the two approaches described above. Average loss and  accuracy.}
        \label{tab:remove_client_3clients_10clients}
    \end{table}

In the previous table we can see that, when there are intermittent clients, the results are better with 3 clients than with 10, both in terms of loss and accuracy, and it is also observed that in both cases, in this example is more interesting to apply the first approach, which is not to keep the last weights calculated for the eliminated client.

Finally, regarding the second problem presented previously related with intermittent clients, the case where a client does not send its weights in time at the end of a repetition of the FL scheme but sends them in another round, we propose two possible approaches (A and B) to be followed. In particular we have studied this issue applying the two approaches to an example with the cases of 3 and 10 clients. In this example it is assumed that at the end of the 5th repetition, prior to aggregation, a client $i$ does not send its updates, but these are received by the server at the end of the 10th repetition, also prior to aggregation. It can be seen that approach A achieves better results in terms of loss and accuracy for the case of 3 clients, while for the case of 10 clients, although the best results are obtained in terms of loss with approach B, in terms of accuracy it remains at 0.6250 with both approaches (the model does not converge). We can also observe that although with 10 clients the model does not overcome the accuracy 0.6250, with approach A there is a great reduction of the loss with respect to the case of 3 clients, i.e., although the model does not converge in terms of accuracy in the case of 10 clients, it does so faster than with 3 clients in terms of loss. 

Table~ \ref{tab:intermittent_clients_mean} shows the mean obtained for the test loss and accuracy after analyzing this case considering each of the possible clients as an intermittent client, which verifies the superiority of approach A over approach B in this example for the case of 3 clients, and the opposite for the case of 10 clients (see Tables~\ref{tab:appAB_3clients} and~\ref{tab:appAB_10clients}). This further supports, as mentioned previously, that the most appropriate procedure should be carefully studied in each case under study, as this same example shows how one approach is more convenient than another depending on the split of the data in a greater or lower number of clients.

    \begin{table}[ht]
        \centering
        \begin{tabular}{@{}lcccc@{}}
        \toprule
        \multicolumn{1}{c}{ } & \multicolumn{2}{c}{\textbf{Approach A (test)}} & \multicolumn{2}{c}{\textbf{Approach B (test)}} \\
        \cmidrule{2-5}
        \multicolumn{1}{c}{ } & \textbf{Average} & \textbf{Average} & \textbf{Average} & \textbf{Average} \\
        \multicolumn{1}{c}{ } & \textbf{loss} & \textbf{accuracy} & \textbf{loss} & \textbf{accuracy} \\
        \midrule
        \textit{3 clients} & 7.0646 & 0.6661 & 5.0948 & 0.7345\\
        \textit{10 clients} & 4.7345 & 0.6250 & 5.7186 & 0.6250\\
        \bottomrule
        \end{tabular}
        \caption{Results obtained by eliminating one client and adding a new one with the two approaches described above. Average loss and  accuracy.}
        \label{tab:intermittent_clients_mean}
    \end{table}

\section{Conclusions and future work}
\label{sec:conclusions}

In this paper we have presented the implementation of a complete federated learning algorithm and its application to a practical use case on medical image classification, with special attention to the evolution of results as the number of clients in which the original data are divided increases. Although the classical use of this type of techniques is motivated by the impossibility of working in a centralized way with data from different clients that intend to study the same subject using data of the same characteristics (because they cannot or do not want to share their data with each other or with an external server due to privacy restrictions), we will also study the advantages that this approach can bring over the classical (centralized) one by simulating different clients in a specific use case. 

Specifically, for the analyzed use case, the different clients have been artificially generated from the training data of a public dataset of medical images of chest X-Ray. Specifically, the objective is to classify the images according to whether or not the patient has pneumonia. In particular, the different results obtained by dividing the initial train set into 3 and 10 clients have been studied. Further to the study of the results for the test data set, for the cases of three and ten clients, individual results have been analyzed for each client test set. It is worth noting that in both cases of this use case, for the same runtime, much better results are obtained by considering $N_e = 1$ and $N_r = 10$ instead of $N_e = 10$ and $N_r = 1$. This implies that in this particular example it is more convenient to apply more rounds of the FL scheme than training epochs of the model.

We can thus see that this technique not only gives us the advantages of greater data security (since data does not travel between clients or between client and server), but also, when applied to an scenario in which the centralized approach could be used, federated learning also shows improvements in terms of reduction of computing time and even in accuracy increasing with respect to the classic approach.

In addition, an analysis has been made of the possible steps to follow in the case of the problem of intermittent clients. In particular, we have started by analyzing two approaches in the case of a client leaving the training and a new one entering, distinguishing in the case of the one who leaves the training the results that would be obtained if the weights sent in previous iterations are no longer considered, compared to if they continue to be used in the aggregation. Moreover, we have also discussed a problem that can be very common, among other factors, due to connectivity. This is the case when a client does not send its updates to the server after a predefined time has elapsed but sends it after certain rounds. In the situation that this happens, we must consider whether to continue using the old updates, and include the new one when it is received, or not to include the weights of this client until they are those corresponding to the current repetition of the FL schema. Specifically, in this case for the specific example under study we have been able to conclude that the second approach is more convenient, as shown in Table~\ref{tab:intermittent_clients_mean}.

As for next steps, we foresee many lines of work and research as a continuation of this work, such as the following:

\begin{itemize}
    \item Study of different models, in particular the application to other kind of machine learning models, not only to neural or convolutional networks.
    \item Continue with the exhaustive study of results and possible responses in the case of intermittent clients: that more than one appears and disappears, and that this occurs in several repetitions of the process.
    \item Selection, implementation and use of other aggregation functions according to the case of study. For example, using \textit{cluster federated averaging operator}, with is based on using the algorithm of k-means and use the centroids in order to obtain the new weights (exposed in the documentation about the aggregation operators of \cite{SherpaFL}). Moreover, as mentioned when presenting the aggregation function used in this use case, it is really attractive to try to consider machine learning models to optimize the aggregation.
    \item Analysis of more use cases, in particular with special attention to the non-i.i.d. data case \cite{Su2022}. Note that when the training data of a model with a federated learning architecture is unbalanced in its distribution, as proved in \cite{8988732}, the accuracy may be reduced (see \cite{Su2022}). 
    \item Analysis of the possible attacks that the process may suffer and study of different ways to avoid them or reduce their effects. Some of these are: corruption of models by attackers (i.e. some clients could be attackers modifying the model with corrupted data), attacks on the communication of weights and the extraction of data characteristics from them, etc. One possibility is to implement the algorithm applying Homomorphic Encryption (HE) in the communication of the weights between the clients and the server.
    \item Implementation of a use case comparing different data decentralization architectures, such as the four proposed together with the Parameter Server one in Figure 2 of \cite{Su2022}: all reduce, ring all reduce, gossip and neighbor architecture. 
    \item Introduce federated transfer learning techniques in order to improve the results obtained with the classical approach (see \cite{zhang2022data} where federated transfer learning is applied for machine fault diagnosis, and \cite{yang2019federated}).
    \item Introduce other techniques and architectures for data decentralization such us split and gossip learning (see \cite{vepakomma2018split} and \cite{FL_IEEE}). 
    \item Finally, note that as this was a preliminary study based on simulated clients, it was not necessary to create the client-server communication structure. However, for a real application use case, it is important to emphasize here that such communication must be encrypted. In addition, among other possible tools, we propose the use of \textit{Serf} 
    or 
    \textit{Consul} 
    from \textit{HarshiCorp}\footnote{https://www.hashicorp.com/} for the orchestration of such communication.
\end{itemize}

\noindent \textbf{CRediT authorship contribution statement}

\noindent \textbf{Judith Sáinz-Pardo Díaz}: Conceptualization, Methodology, Software, Formal analysis, Investigation, Visualization, Writing - Original Draft. 
\textbf{Álvaro López García}: Conceptualization, Funding acquisition, Methodology, Software, Supervision, Writing - Original Draft. \\

\noindent \textbf{Declaration of Competing Interest}

\noindent The authors declare that they have no known competing financial interests or personal relationships that could have appeared to influence the work reported in this paper.\\

\noindent \textbf{Acknowledgements}

\noindent First, we would like to thank the reviewers for their comments, which have contributed to improve this work. The authors acknowledge the funding through the European Commission - NextGenerationEU (Regulation EU 2020/2094), through CSIC's Global Health Platform (PTI Salud Global) and the support from the project AI4EOSC ``Artificial Intelligence for the European Open Science Cloud'' that has received funding from the European Union's Horizon Europe research and innovation programme under grant agreement number 101058593. 

\bibliographystyle{elsarticle-num}
\bibliography{federated_learning}

\end{document}